\documentclass[11pt,a4paper]{article}
\usepackage[hyperref]{acl2018}

\usepackage{times}
\usepackage{latexsym}

\usepackage{url}

\usepackage{graphicx}
\usepackage{color}

\usepackage{algpseudocode}
\usepackage{algorithm}

\usepackage{subcaption}

\newcommand{\ignore}[1]{}

\aclfinalcopy 
\def\aclpaperid{618} 

\renewcommand{\vec}[1]{\mathbf{#1}}
\newcommand{\nocomp}{\ensuremath{\mbox{RNNG}_{-\scriptstyle\rm comp}}}

\newcommand{\rnngplusbs}{RNNG$+\mbox{beam~search}$}

\title{Finding Syntax in Human~Encephalography with Beam~Search}

\author{John Hale$^{\spadesuit,\triangle}$  ~ Chris Dyer$^{\spadesuit}$ ~ Adhiguna Kuncoro$^{\spadesuit,\clubsuit}$ ~  Jonathan R. Brennan$^{\diamondsuit}$\\
$^{\spadesuit}$DeepMind, London, UK\\
$^{\clubsuit}$Department of Computer Science, University of Oxford \\
 $^{\diamondsuit}$Department of Linguistics, University of Michigan\\
 $^{\triangle}$Department of Linguistics, Cornell University \\
 {\small \tt \{jthale,cdyer,akuncoro\}@google.com} ~ {\small \tt jobrenn@umich.edu}
}

\date{}

\begin{document}
\maketitle
\begin{abstract}
Recurrent neural network grammars (RNNGs)
are generative~models of $(\mbox{tree},\mbox{string})$~pairs
that rely on neural~networks to evaluate derivational~choices.
Parsing with them using beam~search yields a variety of
incremental complexity~metrics such as word~surprisal and parser~action count.
When used as regressors against human electrophysiological~responses
to naturalistic~text, they derive two amplitude effects:
an early~peak and a P600-like later peak. 
By contrast, a non-syntactic neural~language model yields no reliable~effects.
Model~comparisons attribute the early~peak to syntactic~composition within the RNNG.
This pattern of results recommends the \rnngplusbs\ 
combination as a mechanistic~model of the syntactic~processing
that occurs during normal human language comprehension.
\end{abstract}

\section{Introduction}
Computational~psycholinguistics has
``always been...the thing that computational~linguistics stood the
greatest~chance of providing to humanity''~\cite{J05-4001}.
Within this broad~area, cognitively-plausible parsing~models
are of particular~interest. They are mechanistic computational~models
that, at some level, do the same~task people do in the course of ordinary language~comprehension.
As such, they offer a way to gain~insight into the operation of
the human sentence~processing mechanism~\citep[for a review see][]{hale:ore}.

As \citet{keller:plausible} suggests,
a promising~place to look for such insights
is at the intersection of (a)~incremental~processing,
(b)~broad~coverage, and (c)~neural~signals from the human~brain.

The contribution of the present~paper is situated precisely at this intersection.
It combines a probabilistic generative grammar~\citep[RNNG;][]{dyer-EtAl:2016:N16-1}
with a parsing~procedure that uses this grammar to manage a collection of syntactic~derivations as it advances from one~word to the next~\citetext{\citealp{stern-fried-klein:2017:EMNLP2017}, cf. \citealp{roark:nle}}.
Via well-known complexity~metrics, the intermediate~states of this procedure yield quantitative predictions about language comprehension difficulty.
Juxtaposing these predictions against data from human~encephalography (EEG),
we find that they reliably derive several amplitude~effects
including the P600, which is known to be associated with syntactic~processing~\citep[e.g.][]{osterhout-holcomb92}.

Comparison with language~models based on long short~term memory networks~\cite[LSTM, e.g.][]{lstm,mikolov:diss,graves12}
shows that these effects are specific to the RNNG.
A further~analysis pinpoints one of these effects to RNNGs'
syntactic~composition mechanism. These positive~findings reframe
earlier null~results regarding the syntax-sensitivity of human~processing~\citep{frank:erp}.
They extend work with eyetracking~\citep[e.g.][]{roark-EtAl:2009:EMNLP,demberg13}
and neuroimaging~\citep{Brennan:2016yu,bachrach:diss}
to higher temporal~resolution.\footnote{Magnetoencephalography
also offers high temporal resolution and as such this work fits into
a tradition that includes \citet{wehbe-EtAl:2014:EMNLP2014}, 
\citet{vanschijndeletal:2015:cmcl}, \citet{wingfield17} and \citet{COGS:COGS12445}.}
Perhaps most significantly, they establish a general~correspondence
between a computational~model
and electrophysiological~responses to naturalistic~language.

Following this Introduction, section~\ref{sec:rnngs} presents
recurrent neural network grammars, emphasizing their suitability
for incremental~parsing. Sections~\ref{sec:beam} then reviews
a previously-proposed beam~search procedure for them.
Section~\ref{sec:metrics} goes~on to introduce the novel~application of this procedure
to human~data via incremental complexity~metrics. Section~\ref{sec:eeg}
explains how these theoretical~predictions are specifically brought~to~bear on EEG~data using regression.
Sections~\ref{sec:langmodels} and \ref{sec:results} elaborate~on the
model~comparison mentioned~above and report the results in a way that isolates the operative~element.
Section~\ref{sec:discussion} discusses these results in relation to
established computational~models.
The conclusion, to anticipate section~\ref{sec:conclusion},
is that syntactic~processing can be~found
in naturalistic speech~stimuli if ambiguity~resolution is modeled as beam~search.

\section{Recurrent neural~network grammars for incremental~processing} \label{sec:rnngs}
\begin{figure}
\includegraphics[width=0.5\textwidth]{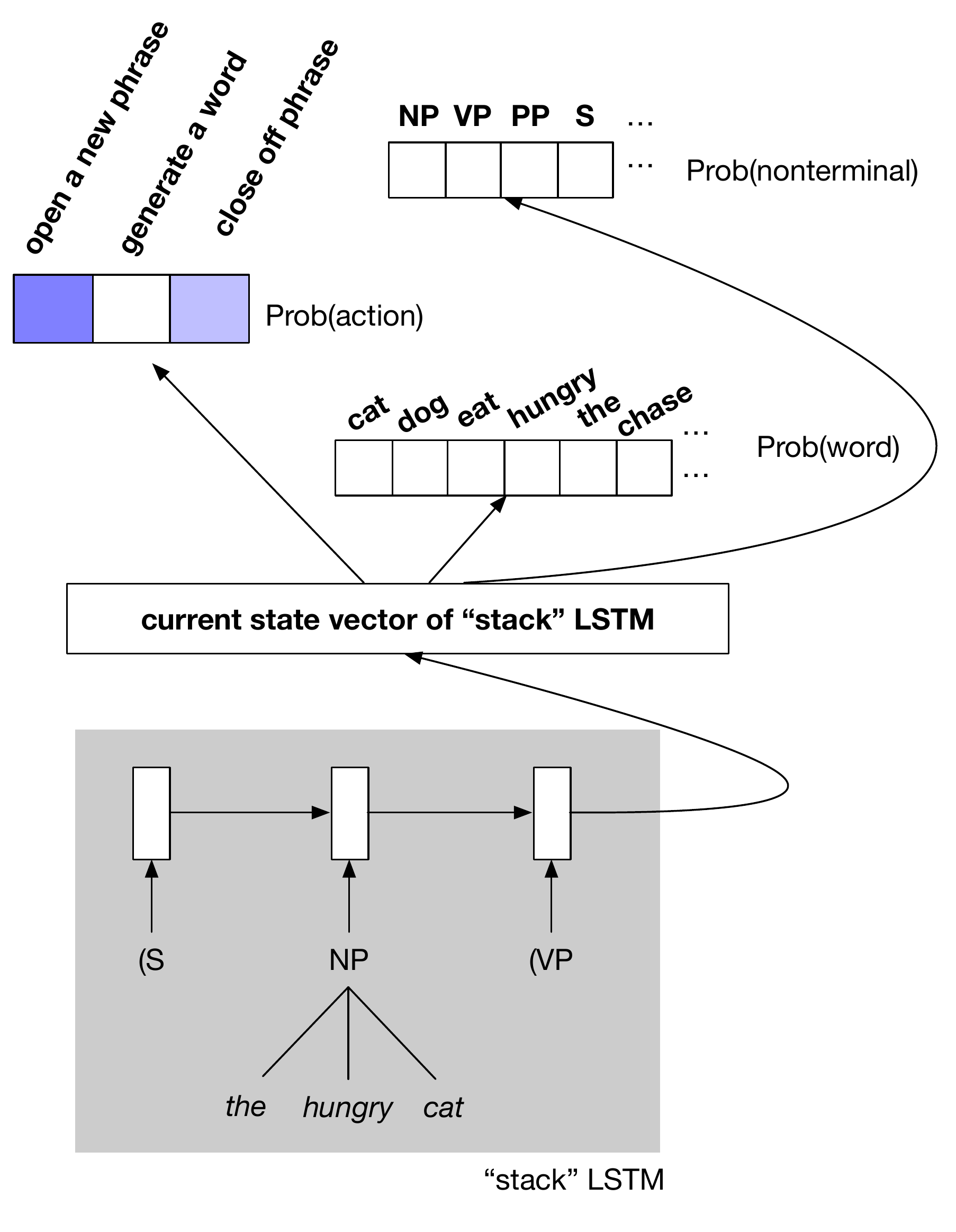}
\caption{Recurrent neural~network grammar configuration used in this paper. The absence of a lookahead~buffer
is significant, because it forces parsing to be incremental. Completed~constituents such as $\left[_{\rm NP}\,\mbox{the hungry cat}\,\right]$
are represented on the stack by numerical~vectors that are the output of the syntactic~composition function depicted in Figure~\ref{fig:composition}.} \label{fig:rnng}
\end{figure}

Recurrent neural~network grammars \citetext{henceforth: RNNGs~\citealp{E17-1117,dyer-EtAl:2016:N16-1}}
are probabilistic~models that generate trees. The probability of a tree
is decomposed via the chain~rule
in~terms~of derivational action-probabilities that are conditioned upon previous~actions
i.e. they are history-based grammars~\citep{black:hbg}.
In the vanilla~version of RNNG, these steps follow a depth-first traversal of the developing phrase structure tree.
This entails that daughters are announced bottom-up one by one
as they are completed, rather than being predicted at the same time as the mother.

\begin{figure}[t]
    \includegraphics[width=0.5\textwidth]{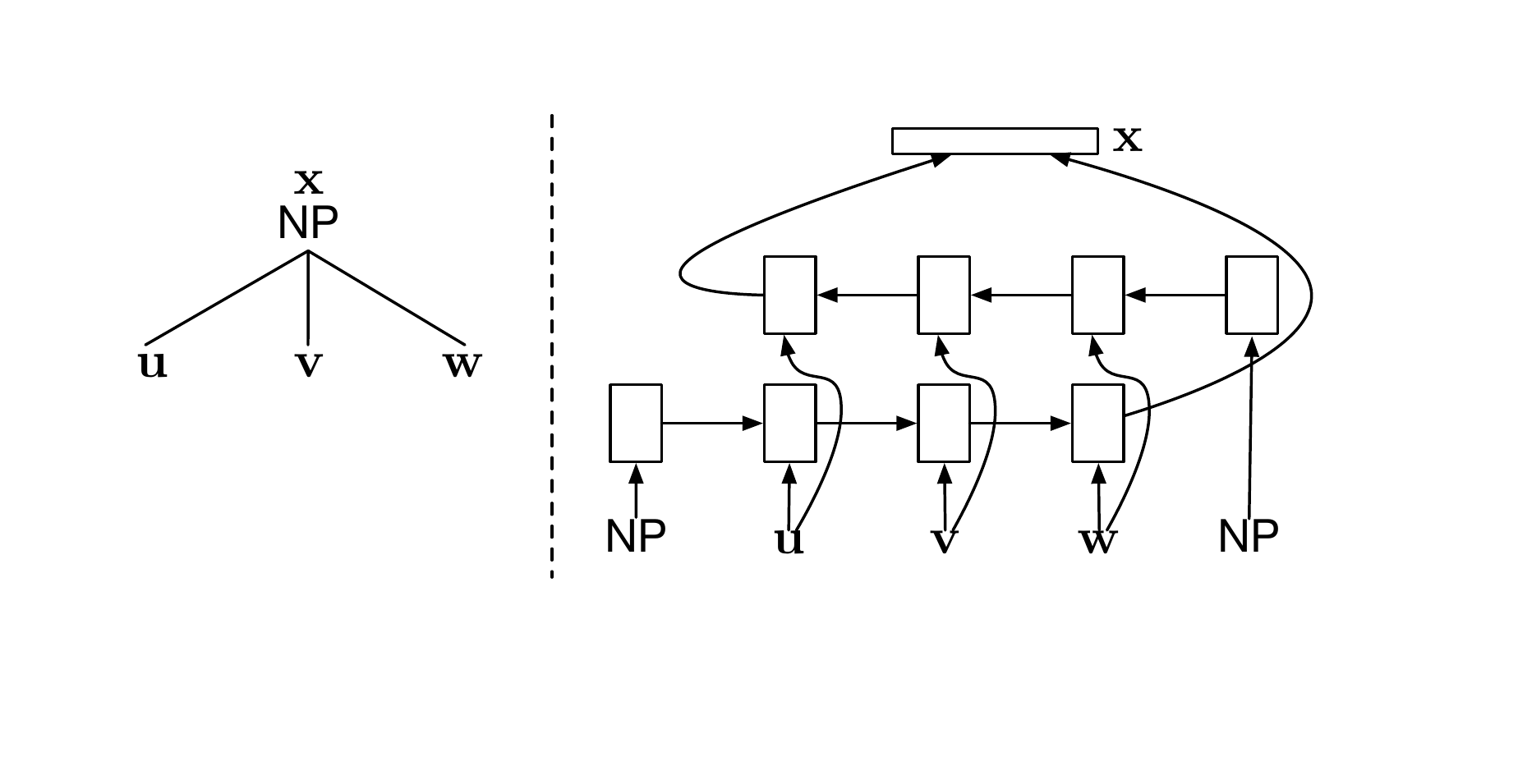}
    \caption{RNNG composition function traverses daughter embeddings~$\vec{u}$, $\vec{v}$ and $\vec{w}$, representing the entire tree with a single vector~$\vec{x}$. This Figure is reproduced from \citep{dyer-EtAl:2016:N16-1}.
} \label{fig:composition}
\end{figure}

Each step of this generative~story depends on the state of a stack, depicted inside the gray~box
in Figure~\ref{fig:rnng}. This stack is ``neuralized'' such~that each stack~entry corresponds to a numerical~vector.
At each stage of derivation, a single vector summarizing the entire stack
is available in the form of the final~state of a neural sequence~model.
This is implemented using the stack~LSTMs of \citet{dyer:stacklstm}.
These stack-summary vectors (central rectangle in Figure~\ref{fig:rnng}) allow RNNGs to be sensitive
to aspects of the left~context that would be~masked by independence~assumptions in a probabilistic context-free~grammar.
In the present~paper, these stack-summaries serve as input to a multi-layer~perceptron
whose output is converted via softmax into a categorical~distribution over three possible~parser actions:
open a new~constituent, close off the latest~constituent, or generate a word.
A hard~decision is made, and if the first or last option is selected,
then the same vector-valued stack--summary is again used, via multilayer~perceptrons,
to decide which specific nonterminal to open, or which specific~word to generate. 

Phrase-closing actions trigger a syntactic~composition function (depicted in Figure~\ref{fig:composition})
which squeezes a sequence of subtree~vectors into one single~vector.
This happens by applying  a bidirectional~LSTM to the list of daughter vectors,
prepended with the vector for the mother~category following \S4.1 of \citet{dyer-EtAl:2016:N16-1}.

The parameters of all these components are adaptively adjusted using backpropagation at training~time, minimizing the cross~entropy relative to a corpus of trees.
At testing~time, we parse incrementally using beam~search as described below in section~\ref{sec:beam}.

\section{Word-synchronous beam search} \label{sec:beam}
Beam~search is one~way of addressing the search~problem that arises with generative~grammars ---
constructive accounts of language that are sometimes said to ``strongly~generate'' sentences.
Strong~generation in this sense simply~means that they derive both an observable word-string as well as a hidden~tree structure.
Probabilistic~grammars are joint~models of these two aspects.
By contrast, parsers are programs intended to infer a good~tree from a given word-string.
In incremental~parsing with history-based models this inference~task is particularly challenging,
because a decision that looks wise at one~point may end up looking foolish in light of future~words.
Beam~search addresses this challenge by retaining a collection called the ``beam'' of parser~states at each word.
These states are rated by a score that is related to the probability of a partial~derivation, allowing an incremental~parser to hedge~its~bets
against temporary~ambiguity. If the score of one~analysis suddenly plummets after seeing some~word, there may still be others within the beam that are not so drastically affected.
This idea of ranked~parallelism has become~central in psycholinguistic~modeling~\citep[see e.g.][]{gibson91,narayanan:bayesian,boston11}.

As \citet{stern-fried-klein:2017:EMNLP2017} observe, the most straightforward application of beam~search
to generative~models like RNNG does not perform well. This is because lexical~actions,  
which advance the analysis onwards to successive~words,
are assigned such low~probabilities compared to structural~actions which do~not advance to the next~word.
This imbalance is inevitable in a probability~model that strongly~generates sentences,
and it causes naive beam-searchers to get bogged~down, proposing more and more phrase~structure
rather than moving on through the sentence. To address it, \citet{stern-fried-klein:2017:EMNLP2017} propose a word-synchronous variant of beam~search.
This variant keeps searching through structural~actions until ``enough'' high-scoring parser~states
finally take a lexical~action, arriving in synchrony at the next~word of the sentence.
Their~procedure is written~out as Algorithm~\ref{alg:wordsynchronous}.

\begin{table*}[]
\begin{tabular}{llllllll}
               & $k$=100 & $k$=200 & $k$=400 & $k$=600 & $k$=800 & $k$=1000 & $k$=2000 \\
\parbox{0.3\textwidth}{\citet{fried-stern-klein:2017:Short} RNNG \\ \hspace*{\parindent} ppl unknown, $-$fast track} & 74.1 & 80.1 & 85.3 & 87.5 & 88.7 & 89.6 & not reported \\
this paper ppl=141, $-$fast track & 71.5 & 78.81 & 84.15 & 86.42 & 87.34 & 88.16 & 89.81 \\
this paper ppl=141, $k_{\textit{ft}} = k/100$ & 87.1 & 88.96 & 90.48 & 90.64 & 90.84 & 90.96 & 91.25 \\
\end{tabular}
\caption{Penn Treebank development section bracketing accuracies (F1) under Word-Synchronous beam~search.
These figures~show that an incremental~parser for RNNG can perform~well on a standard~benchmark.
``ppl'' indicates the perplexity of over both trees and strings  
for the trained~model on the development~set, averaged over words 
In all cases the word~beam is set to a tenth of the action~beam, i.e. $k_{\scriptstyle\rm word} = k/10$.}
\label{tbl:ptb}
\end{table*}

\begin{algorithm}
\begin{algorithmic}[1]
\State $\textit{thisword} \gets$ input beam
\State $\textit{nextword} \gets \emptyset$
\While{$|\textit{nextword}| < k$}
    \State{$\textit{fringe} \gets$ \parbox[t]{0.25\textwidth}{successors of all states $s\in \textit{thisword}$ via any \\ parsing action}}
    \State{prune $\textit{fringe}$ to top $k$}
    \State $\textit{thisword} \gets \emptyset$
    \For{each parser state $s\in \textit{ fringe}$}
        \If{$s$ came via a lexical action}
            \State add $s$ to $\textit{nextword}$
        \Else \Comment{must have been structural}
            \State{add $s$ to $\textit{thisword}$}
        \EndIf
    \EndFor
\EndWhile \\
\Return $\textit{nextword}$ pruned to top $k_{\scriptstyle\rm word} \ll k$
\end{algorithmic}
\caption{Word-synchronous beam search with fast-tracking. After \citet{stern-fried-klein:2017:EMNLP2017}} \label{alg:wordsynchronous}
\end{algorithm}

In Algorithm~\ref{alg:wordsynchronous} the beam is held in a set-valued variable called~\textit{nextword}. 
Beam~search continues until this set's cardinality exceeds the designated action~beam size, $k$.
If the beam~still isn't large enough (line~3) then the search process explores one more action by going around the while-loop again.
Each time through the loop, lexical~actions compete against structural~actions for a place among the top~$k$ (line 5).
The imbalance mentioned above makes this competition fierce, and on many loop iterations \textit{nextword} may not grow by much.
Once there are enough parser~states, another threshold called the word~beam~$k_{\scriptstyle\rm word}$ kicks~in (line 15).
This other threshold sets the number of analyses that are handed~off to the next~invocation
of the algorithm. In the study reported~here the word~beam remains at the default~setting suggested by
Stern and colleagues,~$k/10$.
\vfill

\citet{stern-fried-klein:2017:EMNLP2017} go on to offer a~modification of the basic~procedure called ``fast tracking''
which improves performance, particularly when the action~beam~$k$ is small. Under fast~tracking, an additional step is added between lines 4 and 5 of Algorithm~\ref{alg:wordsynchronous} such that some small number~$k_{ft}$ of parser~states are promoted directly into $\textit{nextword}$. These states are required to come via a lexical~action, but in the absence of fast~tracking they quite possibly would have failed the thresholding step in line~5.

Table~\ref{tbl:ptb} shows Penn~Treebank accuracies for this word-synchronous beam~search procedure, as applied to RNNG.
As expected, accuracy goes~up as the parser considers more and more analyses.
Above $k=200$, the \rnngplusbs\ combination outperforms a
conditional~model based on greedy~decoding (88.9). This demonstration re-emphasizes
the point, made by \citet{brants00} among~others,
that cognitively-plausible incremental~processing can be achieved without loss of parsing performance. 

\section{Complexity~metrics} \label{sec:metrics}

In order to relate computational~models to measured human~responses, some sort of auxiliary~hypothesis or linking~rule is required.
In the domain of language, these are traditionally referred~to as complexity~metrics because of the way they quantify the
``processing complexity'' of particular~sentences. When a~metric offers a prediction on each successive~word,
it is an \emph{incremental} complexity~metric.

Table~\ref{tbl:metrics} characterizes four incremental complexity~metrics that are all obtained
from intermediate~states of Algorithm~\ref{alg:wordsynchronous}.
The metric denoted~\textsc{distance} is the most classic; it is inspired by the count of ``transitions made or attempted'' in \citet{kaplan72}.
It quantifies syntactic~work by counting the number of parser~actions explored by Algorithm~\ref{alg:wordsynchronous} between each word i.e. the number of times around the while-loop on line~3.
The information~theoretical quantities~\textsc{surprisal} and \textsc{entropy}
came into more widespread~use later. They quantify unexpectedness
and uncertainty, respectively, about alternative syntactic~analyses at a given point within a sentence.
\citet{hale16:review} reviews their applicability across many different languages,
psycholinguistic measurement~techniques and grammatical~models.
Recent~work proposes possible relationships between these two metrics, at the empirical as well as theoretical level \citep{vanschijndelschuler:2017:cogsci,pyeongwhan:cmcl18}.

\begin{table}[h!]
\begin{tabular}{lp{0.3\textwidth}}
metric &  characterization \\ \hline
\textsc{distance}  & count of actions required to synchronize $k$ analyses at the next word \\
\textsc{surprisal} & log-ratio of summed forward~probabilities for analyses in the beam\\
\textsc{entropy} & average uncertainty of analyses in the beam \\
\textsc{entropy~$\Delta$} & difference between previous and current entropy value
\end{tabular}
\caption{Complexity Metrics} \label{tbl:metrics}
\end{table}

The \textsc{surprisal}~metric was computed over the word~beam i.e. the $k_{\scriptstyle\rm word}$
highest-scoring partial syntactic~analyses at each successive~word.
In an attempt to obtain a more faithful~estimate, \textsc{entropy} and its first-difference
are computed over \textit{nextword}~itself, whose size varies but is typically much larger than $k_{\scriptstyle\rm word}$.

\section{Regression~models of naturalistic~EEG} \label{sec:eeg} 
Electroencephalography (EEG) is an experimental~technique
that measures very small voltage~fluctuations on the scalp.
For a review emphasizing its implications vis-\'{a}-vis computational~models, see \citet{murphy_wehbe_fyshe_2018}.




We analyzed EEG~recordings from 33~participants as they passively listened
to a spoken~recitation of the first~chapter of \underline{Alice's Adventures in Wonderland}.%
\footnote{\mbox{\url{https://tinyurl.com/alicedata}}}
This auditory stimulus was delivered via earphones in an isolated~booth.
All participants scored significantly better than chance on a post-session 8-question comprehension quiz.
An additional ten datasets were excluded for not meeting this behavioral~criterion, 
six due to excessive~noise, 
and three due to experimenter~error.
All participants provided written informed consent under the oversight of the University of Michigan HSBS Institutional Review Board (\#HUM00081060) and were compensated \$15/h.%
\footnote{A separate~analysis of these~data appears in \citet{brennan18:plos1}; datasets are available from JRB.}

Data were recorded at 500~Hz from 61~active~electrodes 
(impedences $< 25~\textrm{k}\Omega$)
and divided into 2129~epochs, spanning -0.3--1~s
around the onset of each~word in the story.
Ocular artifacts were removed using ICA,
and remaining epochs with excessive noise were excluded.
The data were filtered from 0.5--40 Hz, 
baseline corrected against a 100~ms pre-word interval, and
separated into epochs for content~words and epochs
for function~words because of interactions between parsing variables of interest
and word-class \citep{roark-EtAl:2009:EMNLP}.

Linear~regression was used per-participant, at each time-point and electrode,
to identify content-word EEG~amplitudes that correlate with
complexity~metrics derived from the \rnngplusbs\ combination
via the complexity~metrics in Table~\ref{tbl:metrics}.
We refer to these time~series as Target predictors.

Each Target predictor was included in its own model, along with several Control~predictors
that are known to influence sentence~processing:
sentence~order, word-order in sentence, log word frequency~\citep{Lund1996}, 
frequency of the previous and subsequent word, and acoustic sound~power averaged over the first
50~ms of the epoch. 

All predictors were mean-centered. 
We also constructed null regression models in which the
rows of the design~matrix were randomly permuted.%
\footnote{Temporal auto-correlation across epochs could impact model fits. Content-words are spaced 1 s apart on average and a spot-check of the residuals from these linear models indicates that they do not show temporal auto-correlation:  AR(1) $< 0.1$ across subjects, time-points, and electrodes.}
$\beta$~coefficients for each effect were tested against these null~models at the group~level
across all~electrodes from 0--1~seconds post-onset, using a non-parametric cluster-based permutation~test to correct for multiple~comparisons across electrodes and time-points \cite{Maris:2007lr}.

\section{Language models for literary~stimuli} \label{sec:langmodels}

We compare the fit against EEG data for models that are trained on the same amount of textual~data but
differ in the explicitness of their syntactic~representations.

At the low~end of this~scale is the LSTM language~model. Models of this type treat sentences as a sequence of words,
leaving it up to backpropagation to decide whether or not to encode syntactic~properties in a learned history vector~\citep{linzen2016assessing}.
We use \textsc{surprisal} from the LSTM as a baseline.

RNNGs are higher on this~scale because they explicitly build a phrase~structure tree using a symbolic~stack.
We consider as~well a degraded~version, \nocomp\
which lacks the composition~mechanism
shown in Figure~\ref{fig:composition}. This degraded~version replaces the stack with
initial~substrings of bracket~expressions, following \citet{choe-charniak:2016:EMNLP2016,vinyals2015grammar}.
An example would be the length~$7$ string shown below  
\begin{center}
\begin{tabular}{c|c|c|c|c|c|c}
(S  &  (NP  & the & hungry & cat & )$_{NP}$ & (VP
\end{tabular}
\end{center}
Here, vertical lines separate symbols whose vector encoding would be considered separately by \nocomp .
In this degraded~representation, the noun~phrase is not composed explicitly. It takes~up five~symbols
rather than one. 
The balanced parentheses (NP and )$_{\scriptstyle\rm NP}$ are rather like instructions for some subsequent~agent
who might later perform the kind of syntactic~composition that occurs on-line in RNNGs, albeit in an implicit manner.  

In all cases, these language~models were trained on chapters 2--12 of \underline{Alice's Adventures in Wonderland}.
This comprises 24941 words. The stimulus that participants saw during EEG data~collection,
for which the metrics in Table~\ref{tbl:metrics} are calculated, 
was chapter~1 of the same~book, comprising 2169~words.

RNNGs were trained to match the output~trees provided by the Stanford parser \citep{klein-manning:2003:ACL}.
These trees conform to the Penn~Treebank annotation~standard but do~not explicitly~mark long-distance
dependency or include any empty~categories. They seem to adequately~represent basic syntactic properties
such as clausal~embedding and direct~objecthood; nevertheless we did~not undertake any manual correction.

During RNNG training, the first~chapter was used as a development~set,
proceeding until the per-word perplexity over all parser~actions on this set reached a minimum, 180.
This performance was obtained with a~RNNG whose state~vector was 170~units wide. 
The corresponding LSTM language model state~vector had 256~units; it reached a per-word perplexity of 90.2.
Of course the RNNG estimates the joint probability of both trees \emph{and} words, so these two perplexity levels are not directly~comparable.
Hyperparameter~settings were determined by grid~search in a region near the one which yielded good~performance
on the Penn~Treebank benchmark reported on Table~\ref{tbl:ptb}.

\section{Results} \label{sec:results}
To explore the suitability of the RNNG $+$ beam~search combination as a cognitive~model of language~processing difficulty,
we fitted regression~models as described above in section~\ref{sec:eeg} for each of the metrics in Table~\ref{tbl:metrics}.
We considered six beam~sizes $k=\left\{100,200,400,600,800,1000\right\}$. 
Table~\ref{tbl:significance} summarizes statistical~significance levels
reached by these Target predictors; no other combinations reached statistical~significance.

\begin{table}[h]
\begin{tabular}{lll}
LSTM & \multicolumn{2}{c}{not significant} \\
\textsc{surprisal} & $k=100$ & $p_{cluster} = 0.027$ \\
\textsc{distance} & $k=200$ &  $p_{cluster} = 0.012$ \\
\textsc{surprisal}& $k=200$ & $p_{cluster} = 0.003$ \\
\textsc{distance} & $k=400$ &  $p_{cluster} = 0.002$ \\
\textsc{surprisal} & $k=400$ & $p_{cluster} = 0.049$ \\
\textsc{entropy~$\Delta$} & $k=400$ & $p_{cluster} = 0.026$ \\
\textsc{distance} &$k=600$ & $p_{cluster} = 0.012$ \\
\textsc{entropy} & $k=600$  & $p_{cluster} = 0.014$
\end{tabular}
\caption{Statistical significance of fitted Target predictors in Whole-Head analysis. $p_{cluster}$~values are minima for each Target with respect to a Monte~Carlo cluster-based permutation~test \citep{Maris:2007lr}.} \label{tbl:significance}
\end{table}

\begin{figure*}[t!]
\centering

          \begin{subfigure}[b]{0.5\linewidth}
          \centering
          \includegraphics[width=\linewidth]{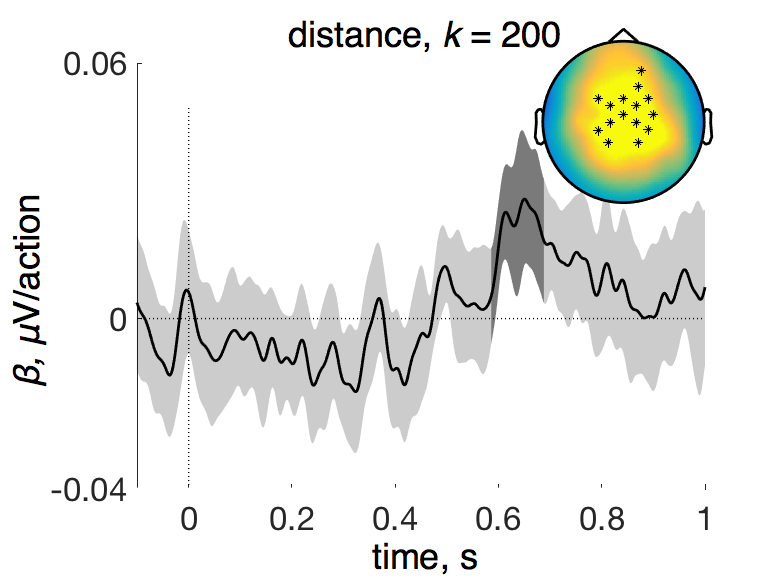}
            \caption{\textsc{distance} derives a P600 at $k=200$.}  \label{fig:p600}
          \end{subfigure}%
          \begin{subfigure}[b]{0.5\linewidth}
          \centering
            \includegraphics[width=\linewidth]{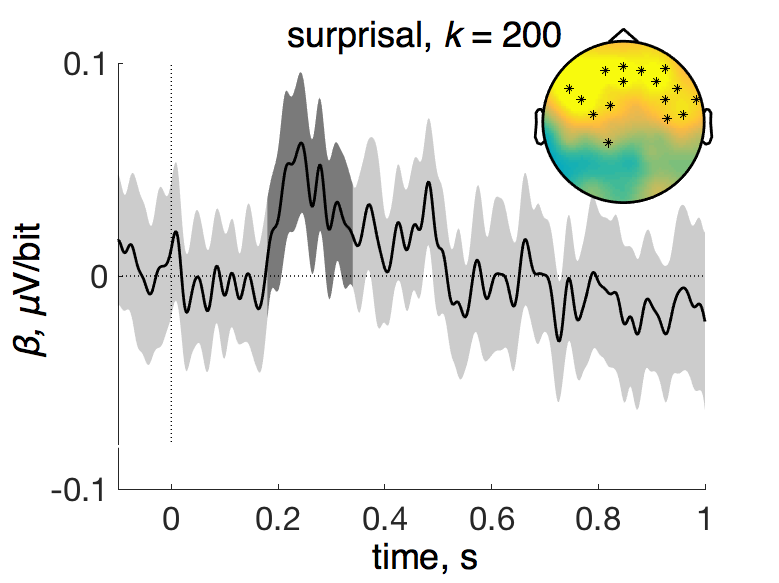}
            \caption{\textsc{surprisal} derives an early~response at $k=200$.}\label{fig:early}
          \end{subfigure}
          \caption{Plotted~values are fitted~regression coefficients and 95\% confidence intervals, statistically~significant in the dark-shaded region with respect to a permutation~test following \citet{Maris:2007lr}. The zero point represents the onset of a spoken~word. Insets show electrodes with significant effects along with grand-averaged coefficient values across the significant time intervals. The diagram averages over all content~words in the first chapter of \underline{Alice's Adventures in Wonderland}.} \label{fig:eegbetas}
\end{figure*}

\subsection{Whole-Head analysis} \label{sec:wholehead}



Surprisal from the LSTM sequence~model did not reliably predict EEG~amplitude at any timepoint or electrode.
The \textsc{distance}~predictor did derive a central~positivity around 600~ms post-word~onset as shown in Figure~\ref{fig:p600}.
\textsc{Surprisal} predicted an early frontal positivity around 250~ms, shown in Figure~\ref{fig:early}. \textsc{Entropy} and \textsc{entropy~$\Delta$} seemed to drive effects that were similarly~early and frontal, although negative-going (not depicted); the effect for \textsc{entropy~$\Delta$} localized to just the left~side.

\subsection{Region of Interest analysis} \label{sec:roianalysis}
\begin{figure}[h]
  \centering
  \includegraphics[width=0.5\textwidth]{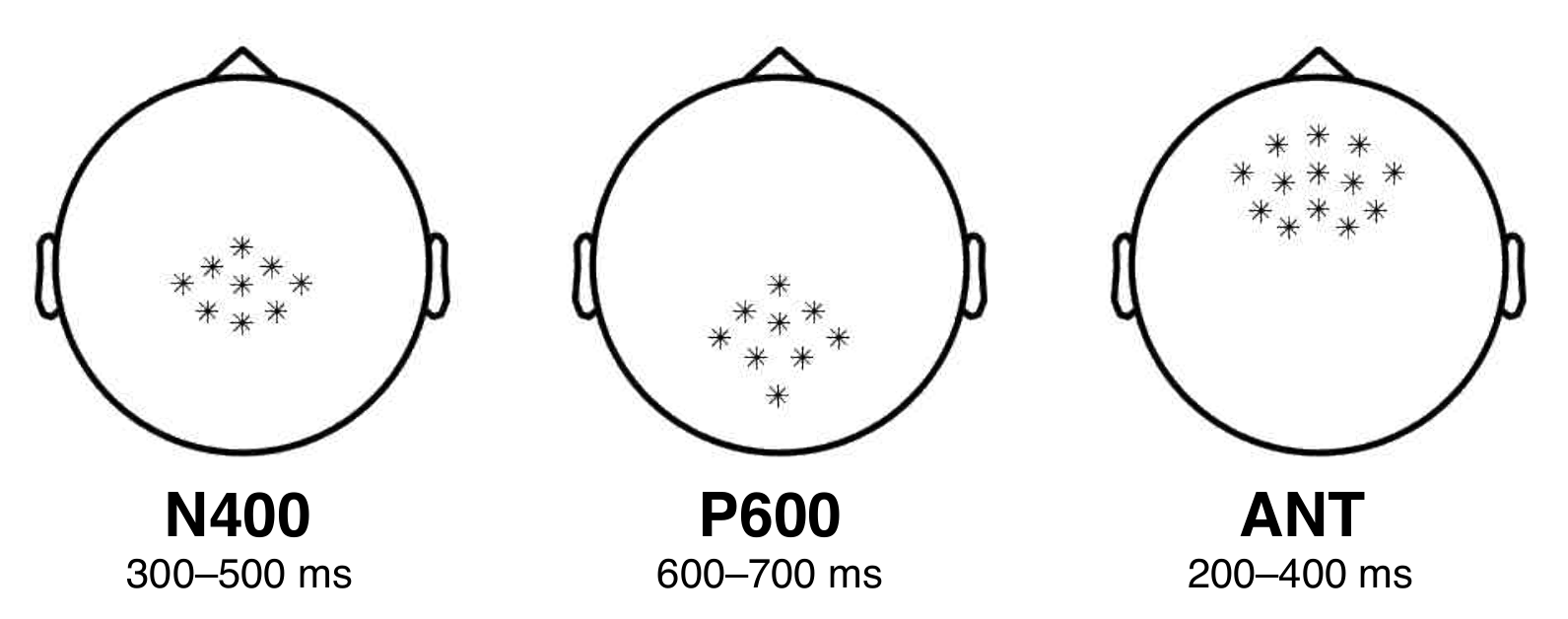}
  \caption{Regions of interest. The first region on the left, named ``N400'', comprises central-posterior electrodes during a time window 300--500~ms post-onset. The middle region,  ``P600'' includes posterior electrodes 600--700~ms post-onset. The rightmost region ``ANT'' consists of just anterior electrodes 200-400~ms post-onset. } \label{fig:rois}
\end{figure}

We compared RNNG to its degraded cousin, \nocomp,
in three regions~of~interest shown in Figure~\ref{fig:rois}.
These regions are defined by a selection of electrodes and a time~window
whose zero-point corresponds to the onset of the spoken~word in the naturalistic speech~stimulus.
Regions ``N400'' and ``P600'' are well-known in EEG~research,
while ``ANT''  is motivated~by findings with a PCFG~baseline reported by~\citet{brennan18:plos1}.

Single-trial data were averaged across electrodes and time-points within each region
and fit with a linear mixed-effects model
with fixed effects as described below and random intercepts by-subjects \citep{AldayENEURO}.
We~used a step-wise likelihood-ratio test to evaluate whether individual Target~predictors from the RNNG
significantly improved over \nocomp, 
and whether a \nocomp\ model significantly improve a baseline regression model. 
The baseline regression model, denoted~$\emptyset$, contains the Control predictors described in section \ref{sec:eeg} 
and \textsc{surprisal} from the LSTM sequence model.
Targets represent each of the eight reliable whole-head effects
detailed in Table \ref{tbl:significance}.
These 24 tests (eight effects by three regions) motivate a Bonferroni correction of $\alpha = 0.002 = 0.05/24$.

Statistically~significant results obtained for \textsc{distance} from \nocomp~in the P600 region 
and for \textsc{surprisal} for RNNG in the ANT region.
No significant results were observed in the N400 region.
These results are detailed in Table~\ref{tbl:neutered}.

\begin{table*}[]
    \centering

    \begin{tabular}{rccccccc}
     & \multicolumn{3}{c}{\nocomp\ $>$  $\emptyset$} & & \multicolumn{3}{c}{RNNG $>$ \nocomp\ } \\
                                         & $\chi^{2}$ & df & $p$ & & $\chi^{2}$ & df & $p$ \\
                                         \hline
    \textsc{distance}, ``P600'' region   &  &  & & &  &  & \\
    $k=200$   & 13.409 & 1 & \textbf{0.00025}  & & 4.198  & 1 & 0.04047 \\
    $k=400$   & 15.842 & 1 & \textbf{$<$0.0001} & & 3.853  & 1 & 0.04966 \\
    $k=600$   & 13.955 & 1 & \textbf{0.00019} & & 3.371  & 1 & 0.06635 \\
    \textsc{surprisal}, ``ANT'' region    &  &  & & &  &  & \\
    $k=100$   & 3.671  & 1 & 0.05537 & & 13.167 & 1 & \textbf{0.00028} \\
    $k=200$   & 3.993  & 1 & 0.04570 & & 10.860 & 1 & \textbf{0.00098} \\
    $k=400$   & 3.902  & 1 & 0.04824 & & 10.189 & 1 & \textbf{0.00141} \\
    \textsc{entropy~$\Delta$}, ``ANT'' region &  &  & & &  &  &  \\
    $k=400$   & 10.141 & 1 &  \textbf{0.00145} & & 5.273  & 1 &  0.02165 
    \end{tabular}

        \caption{Likelihood-ratio tests indicate that regression~models with predictors derived from RNNGs~with syntactic~composition (see Figure~\ref{fig:composition})
        do a better~job than their degraded counterparts in accounting for the early~peak in region ``ANT'' (right-hand columns). Similar comparisons in the ``P600''
        region show that the model improves, but the improvement does not~reach the $\alpha = 0.002$~significance~threshold imposed by our Bonferroni~correction (bold-faced text). RNNGs lacking syntactic composition do improve over a baseline model ($\emptyset$) containing lexical predictors and an LSTM baseline (left-hand columns).} \label{tbl:neutered}
\end{table*}

\section{Discussion} \label{sec:discussion}

Since beam~search explores analyses in descending order of probability, \textsc{distance} and \textsc{surprisal} ought to be yoked,
and indeed they are correlated at $r=0.33$ or greater across all of the beam~sizes~$k$ that we considered in this study.
However they are reliably associated with different EEG~effects. \textsc{surprisal} manifests at anterior~electrodes
relatively early. This seems to be a different~effect from that observed by \citet{frank:erp}.
Frank and colleagues relate N400~amplitude to word~surprisals from an Elman-net, analogous to the LSTM sequence~model evaluated in this work.
Their study found no~effects of syntax-based predictors over and above sequential~ones. In particular, no~effects
emerged in the 500--700~ms~window, where one might have expected a P600.  The present~results, by contrast,
show that an explicitly syntactic~model can derive the P600 quite generally via \textsc{distance}. The absence of an N400~effect
in this analysis could be attributable to the choice of electrodes, or perhaps the modality of the stimulus~narrative, i.e. spoken versus read.

The model~comparisons in Table~\ref{tbl:neutered} indicate that the early~peak,
but not the later~one, is attributable to the RNNG's composition~function.
Choe and Charniak's~\citeyearpar{choe-charniak:2016:EMNLP2016}~``parsing as language~modeling''~scheme
potentially could explain the P600-like wave,
but it would not account for the earlier~peak.
This earlier~peak is the one derived by the RNNG under \textsc{surprisal}, but only when the
RNNG includes the composition mechanism depicted in Figure~\ref{fig:composition}.

This pattern of results suggests an approach to the overall modeling~task.
In this approach, both grammar and processing strategy remain the same, and alternative complexity~metrics,
such as \textsc{surprisal} and \textsc{distance}, serve to interpret the unified~model at different~times
or places within the brain.  This inverts the approach of \citet{brouwer17} and \citet{wehbe-EtAl:2014:EMNLP2014}
who interpret different \emph{layers} of the same neural~net using the same complexity~metric. 

\section{Conclusion} \label{sec:conclusion}

Recurrent neural~net grammars indeed learn something about natural~language syntax,
and what they learn corresponds to indices of human language processing~difficulty
that are manifested in electroencephalography.
This correspondence, between computational~model and human electrophysiological~response,
follows from a~system that lacks an initial~stage of purely string-based processing.
Previous~work was ``two-stage" in the sense that the generative~model
served to rerank proposals from a conditional~model~\citep{dyer-EtAl:2016:N16-1}.
If this one-stage model is cognitively~plausible, then its simplicity undercuts arguments for string-based
perceptual~strategies such~as the Noun-Verb-Noun heuristic~\citep[for a textbook presentation see][]{townsend01}.
Perhaps, as \citet{phillips13:beverbook} suggests, these are unnecessary in an adequate cognitive~model.
Certainly, the road is now~open for more fine-grained investigations
of the order and timing of individual parsing~operations within the human sentence processing mechanism.

\section*{Acknowledgments}
This material is based upon work supported by the National Science Foundation under Grants No.~1607441 and No.~1607251.
We thank Max Cantor and Rachel Eby for helping with data~collection.

\bibliography{john}
\bibliographystyle{acl_natbib}

\end{document}